\documentclass[10pt,twocolumn,letterpaper]{article}

\usepackage{cvpr}
\usepackage{float}
\usepackage{times}
\usepackage{epsfig}
\usepackage{graphicx}
\usepackage{amsmath}
\usepackage{amssymb}
\usepackage{fancyvrb}
\usepackage{relsize}
\usepackage{tabularx}
\usepackage{multicol}

\usepackage{hyperref}
\hypersetup{breaklinks=true,bookmarks=false}

\cvprfinalcopy 

\def\cvprPaperID{****} 

\begin{document}

\title{Deep Convolutional GANs for Car Image Generation}

\author{Dong Hui (Tony) Kim\\
Stanford University\\
{\tt\small tkim6@stanford.edu}
}

\maketitle

\begin{abstract}
In this paper, we investigate the application of deep convolutional GANs on car image generation. We improve upon the commonly used DCGAN architecture by implementing Wasserstein loss to decrease mode collapse and introducing dropout at the end of the discrimiantor to introduce stochasticity. Furthermore, we introduce convolutional layers at the end of the generator to improve expressiveness and smooth noise. All of these improvements upon the DCGAN architecture comprise our proposal of the novel BoolGAN architecture, which is able to decrease the FID from 195.922 (baseline) to 165.966.
\end{abstract}

\section{Introduction}
Generative adversarial networks (GANs) have recently come to the forefront of computer vision research for their ability to learn complex distributions of data. They were first proposed by Ian Goodfellow et al. \cite{goodfellow2014generative} in 2014 as a framework in which two models (a generator and a discriminator) are simultaneously trained. The generator tries to capture the data distribution and generate fake images matching the distribution as closely as possible, while the discriminator attempts to differentiate between real and fake images. In this paper, we investigate the use of GANs to generate and output images of cars, using random noise and images picked from a car dataset as an input. This problem is technically interesting because despite a large body of work making advances on the study of GANs, the training dynamics of GANs are not completely understood \cite{mescheder}. GANs are notoriously hard to train as they must balance between training the generator and discriminator, and their properties of convergence are hard to define. Since there does not seem to be a wide body of work using GANs on car images, we hope that this paper can lend further insight into the training properties of GANs. Another reason for the relevance of applying GANs to car images is that adversarially generated car images could aid the design of future vehicles, as well as providing a useful benchmark for our ability to create convincing images. 
\bigbreak
Throughout this paper, we will show the results of testing deep convolutional GAN sturctures on a given dataset of car images. Since the ultimate goal is to generate convincing images, our evaluation of our results will involve examining the generated images. Quantitatively, we use the Fréchet Inception Distance (FID) as a measurement of the distance between real images and generated images.

\section{Related Work}
As a direct extension of the original proposal of GANs \cite{goodfellow2014generative}, deep convolutional GANs (DCGANs) were proposed in 2015 \cite{radford2015unsupervised}. As the name implies, the generator and discriminator take on a deep convolutional structure (the structure of the generator can be seen in Figure 1), and the use of transposed convolutions in the generator and the use of convolutions in the discriminator lead to more descriptive modeling.
\begin{figure}[H]
\begin{center}
    \includegraphics[width=\linewidth]{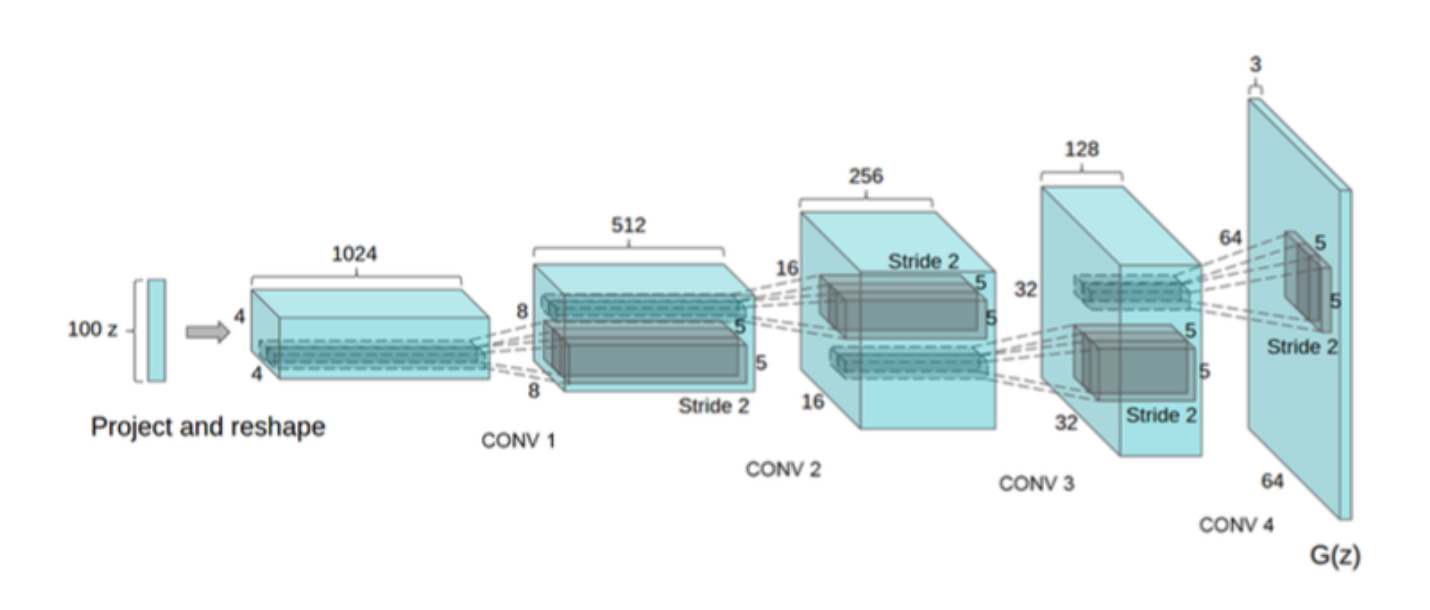}
    \caption{DCGAN Architecture}
\end{center}
\label{fig:DCGAN_arch}
\end{figure}
There are several drawbacks to the raw structure of DCGAN itself, however, and one such drawback is the phenomenon of mode collapse, in which the generated images resemble one another too closely and there is not enough variety in them. There are two approaches that are used to remedy this problem. One such approach is the Unrolled GAN \cite{unrolled}, whereby the generator objective takes into account future versions of the discriminator (or, an ``unrolled optimization'' of the discriminator). The disadvantage to this method is that the computational cost of each training step is directly related to the number of unrolling steps, and there is a tradeoff between more accurate loss values and computational cost. Another approach is the Wasserstein GAN \cite{arjovsky2017wasserstein}, which defines its own loss term to maximize the difference between real and generated images and clips the weights of parameters in the discriminator. The Wasserstein GAN does much to stabilize the training of GANs, but its drawbacks have been associated with the weight clippings, which can lead to a failure to converge or the generation of low-quality samples. Thus, penalties on the norms of the gradients of the critic (or the discriminator) with respect to the input can be used in place of the weight clipping, and this approach is known as WGAN-GP \cite{gp}. However, a problem with both WGAN and WGAN-GP may be that local convergence is not necessarily guaranteed \cite{mescheder}.
\bigbreak
The state of the art models are the original StyleGAN \cite{stylegan} and its second version \cite{stylegan2}, both of which propose a generator structure influenced by style transfer, and Google's BigGAN \cite{biggan}, known for its performance on ImageNet and its application of orthogonal regularization on the generator. The StyleGAN is particularly relevant because, while its paper primarily focuses on human images, it includes a basic application of the architecture to a dataset of cars. While we initially considered using a StyleGAN architecture, it seems that state of the art models such as StyleGAN require more resources than at our disposal.
\bigbreak
Once we were finished with our work, we found a project \cite{asoomar} that had used a similar approach as ours, initially considering a StyleGAN approach but moving onto a DCGAN model when the StyleGAN approach ended up being more computationally expensive than originally predicted. While computationally cheap, the image quality of the generated images were low, pinpointing to deficiencies in the raw DCGAN structure.
\section{Methods}
\begin{figure*}
\centering
    \includegraphics[width=\textwidth]{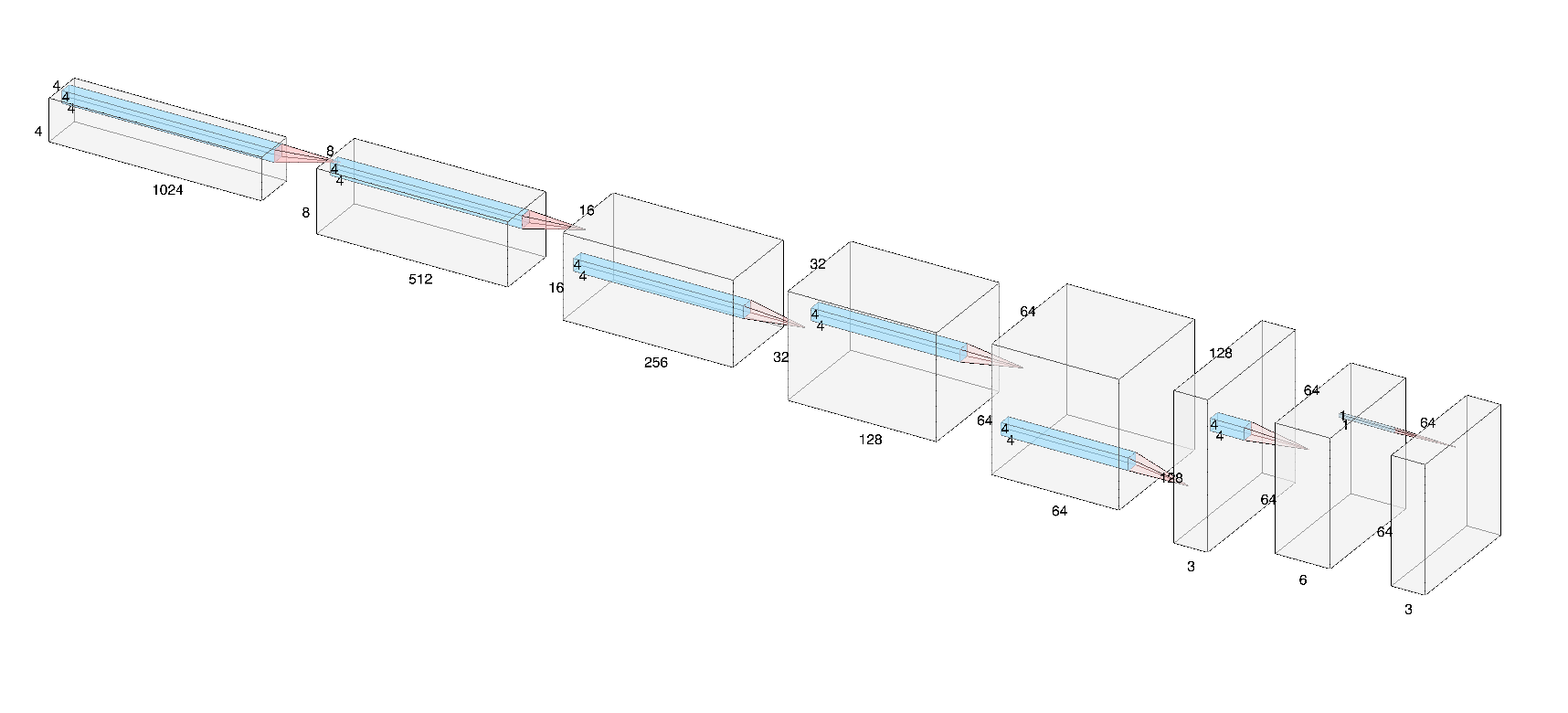}
    \caption{BoolGAN Architecture}
\label{fig:BoolGAN_arch}
\end{figure*}

As our baseline, we used the DCGAN approach outlined by Radford et al. (discussed more extensively in the Related Work section) \cite{radford2015unsupervised}. Our implementation was inspired by a PyTorch tutorial \cite{dcgan_tutorial}. The loss function for DCGAN is defined to be \begin{align*}\underset{G}{\text{min}} \underset{D}{\text{max}}V(D,G) &= \mathbb{E}_{x\sim p_{data}(x)}\big[logD(x)\big] \\ &+ \mathbb{E}_{z\sim p_{z}(z)}\big[log(1-D(G(z)))\big].\end{align*} In other words, for the discriminator, we maximize $log(D(x)) + log(1-D(G(z)))$ by performing gradient ascent, and for the generator, we minimize $log(1-D(G(z)))$ by performing gradient descent. However, this alternation between gradient ascent and descent makes the model hard to learn, so in practice, most maximize $log(D(G(z))$ in the generator by gradient ascent instead, as this brings along performance benefits. Thus, in our implementation, we implemented gradient ascent for both loss terms.
\bigbreak
We then made further improvements on the DCGAN architecture by combining it with the Wasserstein GAN. This in effect rid the discriminator (now called a critic) of its last sigmoid layer, returning scalar scores instead of probabilities. WGAN makes necessary the definition of a 1-Lipschitz function $f$ in lieu of $D(x)$ following the constraint $|f(x_1) - f(x_2)| \le |x_1-x_2|$ so that in the critic we can now maximize $f(x) - f(G(z))$ by performing gradient ascent, and in the generator we can maximize $f(G(z))$. The final component to a Wasserstein to implement was a weight clipping of the parameters in the discriminator, and the exploration of hyperparameter values of the weight clipping is explored in the Experiments and Results section.
\bigbreak
A small improvement we made on top of combining DCGAN and WGAN was the addition of dropout at the end of the discriminator. Because GANs get easily stuck, we thought introducing randomness and stochasticity could help the GAN in those situations and thus improve performance \cite{dropout_explanation}.
\smallbreak
Lastly, on top of all of these improvements, we changed the convolutional architecture of the generator in DCGAN to arrive at our novel BoolGAN architecture as seen in Figure \ref{fig:BoolGAN_arch}. The baseline DCGAN architecture uses a series of transposed 2D convolutional layers (along with 2D batch normalizations and ReLUs) to arrive at a $64\times64$ image with 3 channels. We hypothesized that given the crudeness of DCGANs, there may be a lot of noise in the generated images. Thus, we thought that the addition of a convolutional layer(s) would help smooth such noise and improve the performance of our GAN. This idea was in part inspired by the architecture of Poole et al. \cite{poole} in which $1\times1$ convolutions are added to the end of the generator. To maintain the same $64\times64$ dimensionality returned by the generator, we apply an extra transposed convolutional layer to increase the dimension to $128\times128$ with 3 channels before applying our 2D convolutional layer that changes the dimension to $64\times64$ with 6 channels. Because this is twice the amount of channels as desired, we finally apply a 2D convolutional layer with a $1\times1$ filter that decreases the number of channels from 6 to 3. This is in accordance with the DCGAN formality that there should be no pooling or fully connected layers used. These changes, along with the Wasserstein loss and dropout improvements, comprise the structure of BoolGAN.
\section{Dataset}
We used the dataset provided by Nicolas Gervais, which contains 64,000 images of cars labeled by price, model year, body type, and more \cite{gervais}. All of these images were used as training data, since GANs do not require a validation or testing phase. As a pre-processing step, we resized images (generally of size $320\times210$ with decent resolution) down to $64 \times 64$ before feeding them into our model. Otherwise, there was no data augmentation or other preprocessing done on the images, as the images themselves were in a sense already augmented. As shown in Figure 3, the images in their original state included not only pictures of not only the whole body of cars, but also closeups of various parts of cars (e.g. the A/C controller inside a car, the car hood, etc.). Furthermore, each picture seemed to have been taken with its unique angle, probably due to the fact that multiple sources were used when collecting the dataset. Thus, the dataset as it is fed in a fairly diverse array of images into our models, and we deemed that no further preprocessing was necessary.
\begin{figure}[H]
\begin{center}
  \includegraphics[width=0.9\linewidth]{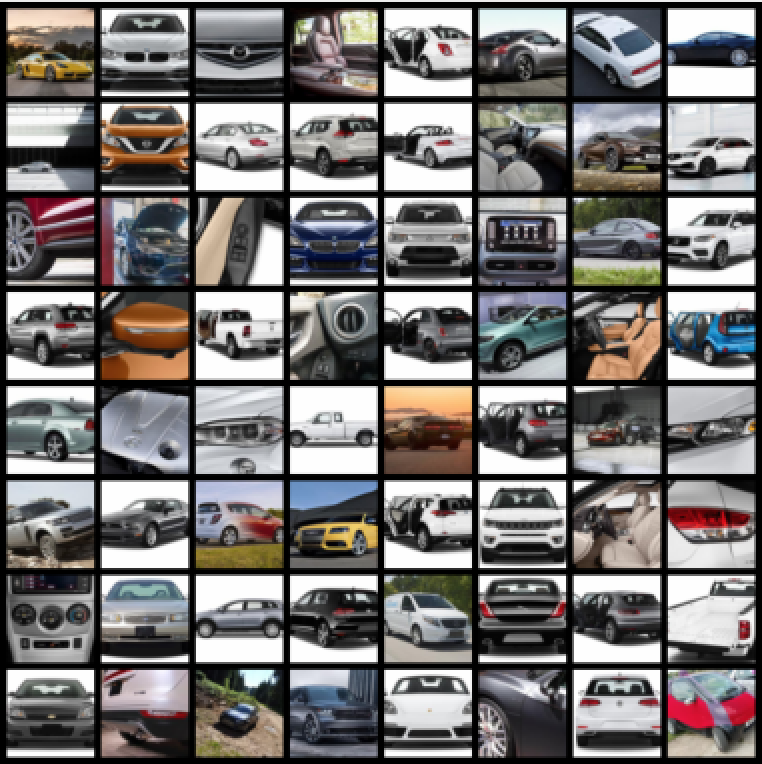}\
  \caption{Real images, sampled from dataset}
\end{center}
\label{fig:real_images}
\end{figure}

\section{Experiments and Results}
We began testing our baseline with manual tuning of hyperparameters (exploring various learning rates and $\beta_1$ values for the Adam optimizer). We investigated with several different learning rates between $1\times10^{-4}$ and $2.5\times10^{-4}$, and we discovered that low learning rates made the graph hard to converge and resulted in less realistic images while the high learning rates led to overshooting and a lot of fluctuation in the loss. We also experimented with higher $\beta_1$ values to give more weight to the cache, but we observed that $\beta_1$ values that were too high were making it hard for the updates to explore in different directions and leading the model to get stuck (possibly at some local minima or saddle point); thus, the loss function was actually observed to increase in this case. In the end, for the baseline, we determined the best results occurred with 50 epochs, a learning rate of $2 \times 10^{-4}$, a $\beta_1$ value of 0.5, batch size equal to 128, and a $\beta_2$ value set to 0.999 (settings that were recommended in Radford et al. \cite{radford2015unsupervised}).
As for the WGAN hyperparameters, we originally tried using the settings recommended by Arjovsky et al. (learning rate = $5\times10^{-5}$, weight clipping constant $c=0.01$, and RMSProp optimizer) \cite{arjovsky2017wasserstein}, but we found that settings closer to that of the baseline worked better (learning rate = $2\times10^{-4}$, weight clipping constant $c=0.1$, and Adam optimizer with $\beta_1=0.5$ and $\beta_2=0.999$). As for dropout, we tried both $p=0.2$ and $p=0.5$ as was commonly recommended, and $p=0.2$ gave better performance.
\bigbreak
Finally, as for BoolGAN's hyperparameters, we kept most settings from the DCGAN+WGAN+dropout, but the learning rate was changed, since the $2\times10^{-4}$ rate prevented the model from making true progress. We had to adjust the learning rate up to $7.5\times10^{-4}$ for better performance, and this may be because the BoolGAN model has more layers through which to propagate its gradient updates. We recommend at least starting the learning rate at $7.5\times10^{-4}$, and learning rates can be decreased incrementally to improve performance.
\bigbreak
As mentioned before, we evaluate our results via the Fréchet Inception Distance (FID). This is a measurement of the distance between two multivariate Gaussians with mean $\mu$ and covariance $\Sigma$. The FID between the distribution of real images $r$ and generated images $g$ is defined as $$\text{FID} = ||\mu_r - \mu_g||^2 + \text{Tr} (\Sigma_r + \Sigma_g - 2 (\Sigma_r \Sigma_g)^{1/2}),$$ where $\text{Tr}$ sums up all the diagonal elements \cite{fid_math}. Thus, a lower FID score means that the distribution of generated images matches the distribution of real images more closely. 
\bigbreak
What follows are the FID scores across all of our tested models.
\begin{figure}[H]
\begin{center}
\begin{tabularx}{.8\linewidth}{|X|c|}
    \hline
    Model & FID Score \\
    \hline
    Baseline DCGAN & 195.922\\
    \hline
    DCGAN + dropout & 183.113\\
    \hline
    DCGAN + WGAN & 179.987\\
    \hline
    DCGAN + WGAN + dropout & 176.031\\
    \hline
    \textbf{BoolGAN} & \textbf{165.966}\\
    \hline
\end{tabularx}
\end{center}
\caption{FID scores of all tested models}
\label{fig:FID}
\end{figure}
As Figure \ref{fig:FID} shows, each improvement to the baseline seems to have improved the FID score, with our proposed BoolGAN performing the best in terms of FID score. All of these models follow a similar pattern of FID score decrease during training time as Figure \ref{fig:FID_graph} shows, with the FID decreasing rapidly at first and slowing its decrease asymptotically. As mentioned before, the exact moment of convergence is hard to identify for GANs, so with the exception of BoolGAN (which trained for longer until the FID graph seemed to become flat), the other models were capped at 50 epochs because it seems that their less sophisticated architectures required less iterations for the FID value to settle.
\begin{figure}[H]
\begin{center}
    \includegraphics[width=\linewidth]{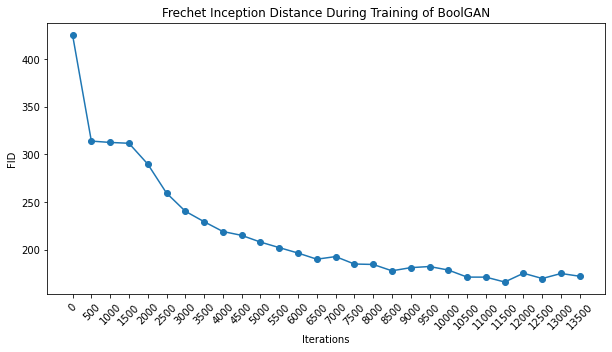}
    \caption{Progress of FID throughout training of BoolGAN}
\end{center}
\label{fig:FID_graph}
\end{figure}

In terms of the quality of images generated, there are differences between the outputs of the baseline DCGAN and the outputs of the improved models (most notably, BoolGAN). Comparing Figures 6 and 7, each of which show 64 examples of generated images from the baseline DCGAN and the BoolGAN, respectively, there are 13 images of easily recognizable cars from the baseline collection (Figure 6) versus 20 from the BoolGAN collection (Figure 7). Furthermore, for images that are not easily recognizable as cars, the baseline DCGAN often outputs images that are very far from being recognizable, often showing swirls of arbitrary colors, whereas the BoolGAN may contain images that seem very close to being cars but are missing a part (for example, the image in the 5th row and 6th column of Figure 7 is very close to being recognizable as a car and only needs wheels).

\begin{figure}[H]
\begin{center}
    \includegraphics[width=\linewidth]{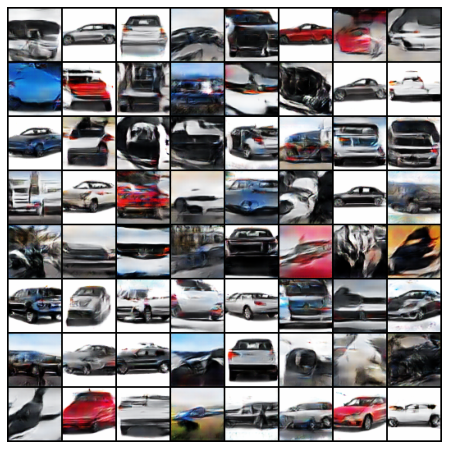}
    \caption{Generated Images of Baseline DCGAN}
\end{center}
\label{fig:baseline_array}
\end{figure}

\begin{figure}[H]
\begin{center}
    \includegraphics[width=\linewidth]{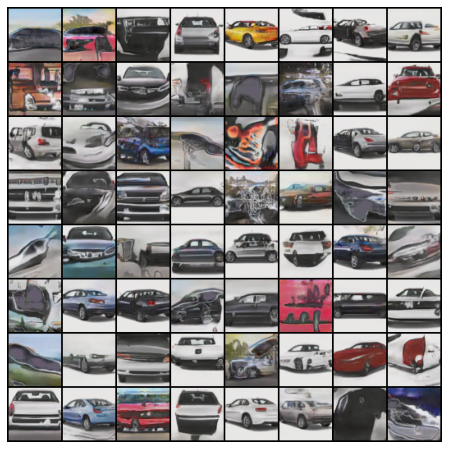}
    \caption{Generated Images of BoolGAN}
\end{center}
\label{fig:boolgan_array}
\end{figure}

Even when comparing the BoolGAN outputs with the outputs of the second best model (DCGAN+WGAN+Dropout) shown in Figure 8, the BoolGAN outputs' image quality is better, with less arbitrary splotches of color on images that can be recognized as cars. For example, in the 6th row of Figure 8, there are splotches of red on the body of the white cars, and it is hard to see that kind of pattern in Figure 7. This may be because the convolutional layers smooth out arbitrary noises that are not wanted. Furthermore, there seems to be a more diverse sample of generated cars from the BoolGAN in Figure 7 as opposed to Figures 6 and 8; of course, while this may be due to pure coincidence, Figure 7 contains a yellow car and has more color variety among recognizable cars as opposed to the two other figures. This may be evidence that BoolGAN is able to decrease mode collapse and output a larger variation of images. Of course, the mode collapse problem may still be present in BoolGAN outputs, as many of the car images in Figure 7, although having different colors, have cars of the same shape and are of the same angle. 

\begin{figure}[H]
\begin{center}
    \includegraphics[width=\linewidth]{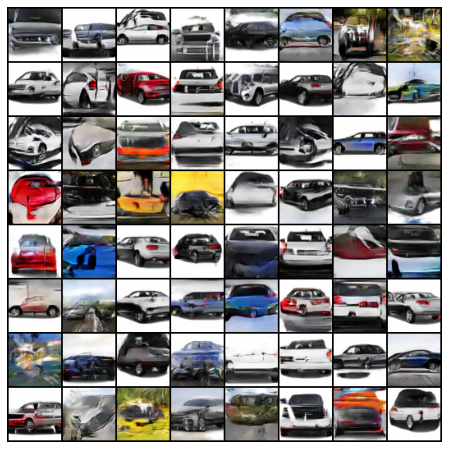}
    \caption{Generated Images of DCGAN+WGAN+Dropout}
\end{center}
\label{fig:wgan_dropout}
\end{figure}

\section{Conclusion and Future Work}
After having tested 5 different deep convolutional GANs on our dataset of cars, it seems that our proposed BoolGAN architecture improves on the baseline DCGAN architecture originally proposed by Radford et al. \cite{radford2015unsupervised}. The addition of the Wasserstein loss led to a decrease in mode collapse and FID score, which may be evidence of increased stability in the training of the GAN. The addition of the dropout layer also led to a decrease in the FID score, showing that the addition of stochasticity and randomness can lead to better performance. Furthermore, the addition of convolutional layers at the end of the generator for the BoolGAN architecture seems to have improved the expressiveness of the model and smoothed noise.
\bigbreak
Given the seeming benefits that the addition of convolutional layers at the end of the generator, more experiments can be conducted on how many convolutional layers can be added and of what size those layers can be. Further research can also be done in terms of searching for more ideal hyperparameters, especially since an incremental decrease throughout training in learning rates seem to have improved performance in the BoolGAN. A more rigorous approach to learning rate scheduling may improve the performance of the model.

\section{Contributions and Acknowledgements}
A lot of the code for the DCGAN implementation is inspired by the PyTorch tutorial \cite{dcgan_tutorial}. I would like to give my special thanks to Evan Mickas who was my project partner at the beginning and suggested the ideas of Wasserstein loss and dropout.

\newpage

{\small
\bibliographystyle{ieee_fullname}
\bibliography{final_paper}
}








\end{document}